%% file: main.tex
\documentclass[10pt,twocolumn,letterpaper]{article}

\usepackage[pagenumbers]{cvpr} 

\input{preamble}

\definecolor{cvprblue}{rgb}{0.21,0.49,0.74}
\usepackage[pagebackref,breaklinks,colorlinks,allcolors=cvprblue]{hyperref}

\usepackage{algorithm}
\usepackage{algpseudocode}
\usepackage{tabularx}
\makeatletter
\newcommand{\multiline}[1]{%
  \begin{tabularx}{\dimexpr\linewidth-\ALG@thistlm}[t]{@{}X@{}}
    #1
  \end{tabularx}
}
\makeatother

\def\paperID{5990} 
\def\confName{CVPR}
\def\confYear{2025}

\title{Instance-wise Supervision-level Optimization in Active Learning}

\author{
Shinnosuke Matsuo$^1$, Riku Togashi$^2$, Ryoma Bise$^1$, Seiichi Uchida$^1$, Masahiro Nomura$^2$ \\
$^1$Kyushu University, Japan \quad 
$^2$CyberAgent, Inc., Japan\\
}

\begin{document}
\maketitle

\begin{abstract}
Active learning (AL) is a label-efficient machine learning paradigm that focuses on selectively annotating high-value instances to maximize learning efficiency. Its effectiveness can be further enhanced by incorporating weak supervision, which uses rough yet cost-effective annotations instead of exact (i.e., full) but expensive annotations. We introduce a novel AL framework, Instance-wise Supervision-Level Optimization (ISO), which not only selects the instances to annotate but also determines their optimal annotation level within a fixed annotation budget. Its optimization criterion leverages the value-to-cost ratio (VCR) of each instance while ensuring diversity among the selected instances.
In classification experiments, ISO consistently outperforms traditional AL methods and surpasses a state-of-the-art AL approach that combines full and weak supervision, achieving higher accuracy at a lower overall cost.
This code is available at https://github.com/matsuo-shinnosuke/ISOAL.
\end{abstract}
\section{Introduction}
\begin{figure*}[t]
    \centering
    \includegraphics[width=.85\linewidth]{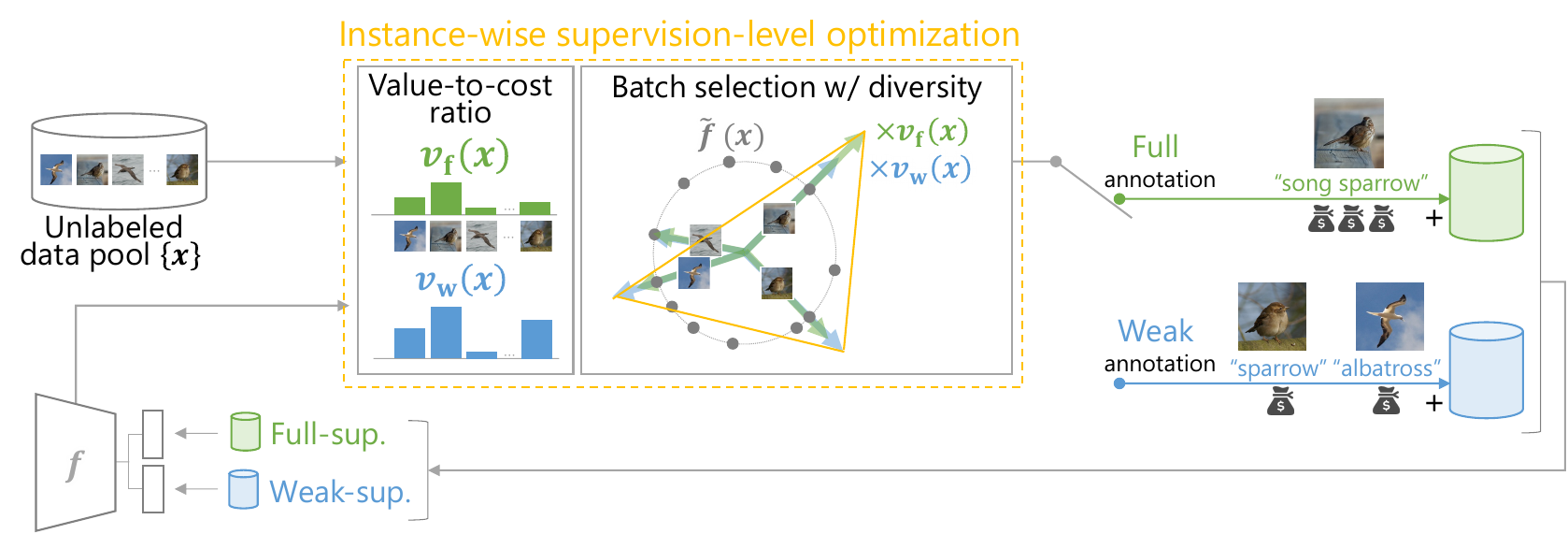}
    \caption{Overview of Instance-wise Supervision-level Optimization (ISO) for active learning. The proposed approach combines weak and full supervision within an active learning framework to maximize annotation efficiency under a fixed budget. First, each instance $\bx$ in the unlabeled data pool is evaluated for its value-to-cost ratio ($\vf(\bx)$ for full supervision and $\vw(\bx)$ for weak supervision). Based on these values and data diversity, instances are selected within the budget constraints for weak supervision (with superclass labels) or full supervision (with exact class labels). This adaptive allocation of supervision-level enables optimal use of resources within the budget constraints in active learning.}
    \label{fig:overview}
\end{figure*}
Machine learning technology has achieved remarkable success in recent years across various fields, including computer vision, natural language processing, and robotics.
However, a major challenge remains: creating labeled datasets for training models often requires significant time and cost, especially in specialized domains.

To address this issue, active learning~\cite{settles2009active,ren2021survey,aggarwal2014active,sener2018active,hacohen2022active,sinha2019variational,ashdeep,kirsch2019batchbald,roth2006margin,wang2014new,houlsby2011bayesian} has been widely studied to improve model performance under limited budgets. 
In active learning, instances (data samples) for annotation are selected iteratively from an unlabeled dataset, followed by annotation and model retraining. 
This repeated cycle improves the model's accuracy while progressively minimizing annotation costs. A common approach in active learning is uncertainty-based sampling~\cite{roth2006margin,wang2014new,houlsby2011bayesian,hsu2015active,ashdeep,tejero2023full,gal2017deep,yoo2019learning}, where instances with high uncertainty, such as low confident of the model's prediction, are selected for annotation, as these are expected to contribute most effectively to model performance.


Another approach to reduce annotation costs is {\em weak supervision}~\cite{zhou2018brief,ren2023weakly,zhang2020survey}. For classification tasks, a weak supervision approach is often designed to just attach rough class labels to individual instances. For example, instead of attaching the exact class label ``song sparrow'' or ``house sparrow'' to an image instance, annotators can attach a rough class label ``sparrows.'' This approach drastically reduces annotation costs (budgets) because the fee paid to a non-expert annotator, who can only assign rough class labels rather than exact ones via a crowdsourcing service, is lower than that of an expert annotator.
Hereafter, we call rough classes as superclasses; therefore, ``sparrows'' is a superclass of the exact class  ``song sparrow.'' 
\par
In general, weak supervision approaches still need instances with {\em full supervision}, that is, instances with the exact class labels. This is simply because it is impossible to determine the class boundary between, ``sparrows'' is a superclass of ``song sparrow'' without the instances with these exact class labels. 
Consequently, we need to balance weak supervision and full supervision under a pre-specified total annotation budget. In other words, we need to optimize the {\em supervision levels} of individual instances. 
Since the total annotation budget is pre-specified as a constraint, the number of instances with full supervision will be limited. More formally, this supervision-level optimization task can be seen as an optimal resource allocation problem with the constraint on available resources.
\par
This paper proposes a label-efficient machine learning approach, called Instance-wise Supervision-level Optimization (ISO) for active learning, as shown in Fig.~\ref{fig:overview}. Roughly speaking, the proposed approach is a sophisticated hybrid of weak supervision and active learning. Different from the state-of-the-art hybrid approach~\cite{tejero2023full} (reviewed in Section 2), our approach has a novel mechanism that automatically determines the optimal supervision levels of individual instances. Specifically, as shown in Fig.~\ref{fig:overview}, all unlabeled instances at a certain round are automatically divided into three categories: instances for full supervision (with exact class annotation), instances for weak supervision (with superclass annotation), and instances that remain unlabeled. The instances selected in the first two categories are considered to be selected for active learning; annotators, therefore, attach appropriate labels to these instances according to their supervision levels.
\par




In our experiments, the proposed method consistently outperformed existing approaches across multiple datasets, demonstrating a more efficient use of the annotation budget. Our approach achieved the same accuracy as the comparative methods while using only three-fifths of the budget by incorporating weak labels when the cost of weak label annotation is half that of fully supervised annotation.
Furthermore, in comparison with the state-of-the-art method that combines active learning and multiple supervision, our approach achieved significantly better cost efficiency by leveraging instance-specific information.

\label{sec:intro}
The main contributions of this paper are summarized as follows: 
\begin{itemize} 
\item We introduce a novel active learning framework that optimizes the supervision level for each instance, allowing for dynamic allocation of weak and full annotations based on budget constraints. 
\item We propose an efficient algorithm that assigns the optimal supervision level to each instance by considering value-to-cost ratios and data diversity, enhancing the cost-effectiveness of the annotation process. 
\item We validate the effectiveness of our approach on classification tasks, using superclasses as weak supervision, and demonstrate its superiority over conventional AL methods and the state-of-the-art AL method that incorporates both full and weak supervision. 
\end{itemize}

\section{Related work\label{sec:related}}
Active learning (AL)~\cite{settles2009active,ren2021survey,aggarwal2014active} is a technique aimed at maximizing model performance under a limited annotation budget by strategically selecting informative instances for annotation.
A widely adopted approach in active learning is uncertainty-based sampling~\cite{roth2006margin,wang2014new,houlsby2011bayesian,hsu2015active,ashdeep,tejero2023full,gal2017deep,yoo2019learning}.
For example, \cite{roth2006margin} selects instances with small margins, prioritizing instances for which the model has low confidence.
Another common approach focuses on maximizing diversity within selected instances~\cite{mccallum1998employing,sener2018active,hacohen2022active,sinha2019variational}, which aims to cover a broad range of the data distribution.
For instance, \cite{sener2018active} selects a representative core set of instances to maximize diversity.
Moreover, there are methods that combine uncertainty and diversity-based strategies to leverage the advantages of both approaches~\cite{ashdeep,kirsch2019batchbald}.
Additionally, some AL methods~\cite{kiyasseh2022soqal,chakraborty2020asking} address the setting where different annotators have varying accuracy levels. These methods aim to jointly optimize both the selection of data for annotation and the choice of annotators to minimize label noise. However, all of these approaches assume a single supervision-level and do not incorporate multiple supervision-levels. 

Another approach to reduce annotation costs is weakly supervised learning (WSL)~\cite{zhou2018brief,ren2023weakly,zhang2020survey}, where less detailed or approximate labels instead of fully supervised labels are utilized. 
In many real-world applications, obtaining fully supervised labels can be expensive and time-consuming. To address this, WSL methods use weak supervision, which is typically easier and cheaper to obtain. For example, in classification tasks, weak supervision can be provided by superclasses such as categories, and in segmentation tasks, it can be provided by bounding boxes, allowing faster labeling with lower costs. 

A hybrid approach between AL and WSL, where the proportion of full and weak supervision is adaptively determined, has been proposed in \cite{tejero2023full}.
Although this approach aims to optimize budget use by setting an ideal balance of weak and full supervision, it applies random sampling after setting the proportion \emph{without} instance-level considerations. 
To our knowledge, no method exists that dynamically assigns full and weak supervision for each instance.

\begin{table}[t]
\caption{Comparison of conventional active learning methods. ``Uncertainty'' (Unc.) refers to whether a method incorporates uncertainty-based instance selection, such as selecting instances with minimal margins. ``Diversity'' (Div.) indicates if the method considers data diversity to maximize the representativeness of selected instances. ``Multiple supervision'' (Mul.) refers to the capability to utilize multiple levels of supervision, such as both weak and full supervision. ``Instance-wise'' (Ins.) represents the ability to choose supervision for each instance adaptively.
\label{tab:RW}}
\vspace{-1mm}
\centering
\scalebox{0.89}{
\begin{tabular}{l|cccc|c}
\hline
                     & Margin,& Coreset, & BADGE, & APFWA & Ours  \\ 
                     & etc.& etc. & etc. & \cite{tejero2023full}&   \\ 
                     & \cite{roth2006margin,wang2014new,houlsby2011bayesian}&\cite{sener2018active,hacohen2022active,sinha2019variational} & \cite{ashdeep,kirsch2019batchbald} & &   \\ \hline
Unc.           & \checkmark    &   $\times$  &   \checkmark            &  $\times$  &  \checkmark     \\
Div.           & $\times$          &  \checkmark & \checkmark   & $\times$   & \checkmark      \\
Mul. & $\times$     &   $\times$ &   $\times$   &  \checkmark      & \checkmark      \\
Ins.        & \checkmark     &  \checkmark & \checkmark   & $\times$    &  \checkmark     \\ \hline
\end{tabular}
}
\end{table}


Table~\ref{tab:RW} summarizes a comparison between popular active learning methods and our instance-wise supervision-level optimization. Each method is evaluated on its ability to incorporate uncertainty, diversity, multiple levels of supervision, and instance-wise adaptability, as defined in the caption. Most uncertainty-based and diversity-based AL methods do not consider multiple levels of supervision, focusing instead on selecting instances based on a single type of annotation. Although APFWA incorporates multiple levels of supervision, it lacks instance-wise optimization. In contrast, our approach is unique in that it optimizes the supervision level on an instance-wise basis, simultaneously considering data uncertainty, diversity, and the annotation cost associated with multiple levels of supervision.

\section{Problem setting \label{sec:problem}}

Let $\Du$ represent the unlabeled data pool, and let $T$ denote the number of rounds in the active learning process, where $B$ is the available budget in each round. 
In this setting, a set of instances is selected to be annotated with either full supervision or weak supervision, depending on the required level of detail and the associated cost.
Specifically, we define $\costf$ as the annotation cost per instance for full supervision, where precise labels are assigned, and $\costw$ as the annotation cost per instance for weak supervision, which involves less precise, lower-cost labeling. These annotation costs, $\costf$ and $\costw$, are provided in advance.

The process of batch selection and training is repeated for $T$ rounds, allowing the model to progressively improve by iteratively selecting and annotating the most informative instances within each budget-constrained round. Once an instance is annotated, it is moved from the unlabeled pool $\Du$ to either the fully supervised data pool $\Df$ or the weakly supervised data pool $\Dw$, depending on the supervision level chosen. 
These annotated instances are then used to train a classifier.

The ultimate goal in this setting is to obtain the classifier's parameters to classify the exact class of input data. To train the classification network, both fully supervised data and weakly supervised data contribute to training the feature extractor in the network. Therefore, balancing fully and weakly supervised data in a cost-effective manner is crucial for optimizing the learning process within budget constraints.

\section{\smash{Proposed method: Instance-wise Supervision}-level Optimization (ISO) \label{sec:method}}
\begin{algorithm}[t]
\caption{ISO: Instance-wise Supervision-level Optimization}
\label{alg:overview}
\begin{algorithmic}[1]

\State \textbf{Inputs: } 
Round $T$, budget of each round $B$, annotation cost $\costf,\costw$, unlabeled data pool $\Du$, feature extractor $f$, classification head $\hf$, $\hw$.
\State \textbf{Outputs: } trained $f^T$, $\hf^T$.

\For {round $t=1, \ldots, T$}
\If {$t$ is $1$}
  \State \multiline{Sample randomly instances from $\Du$ within budget $B$ to create an initial fully supervised dataset $\Df$ and weakly supervised dataset $\Dw$ with an equal number of instances.}
\ElsIf{$t > 1$}
  \State \multiline{Compute the model improvements, $\Mf^t, \Mw^t$, for each supervision using $\Df$ and $\Dw$ by Eq.~(\ref{eq:M1}) and (\ref{eq:M2}).}
  \For {$\bx$ in $\Du$}
  \State \multiline{\textbf{1.} Compute the uncertainties, $\uf^t(\bx),\uf^t(\bx)$, for each supervision.}
  \State \multiline{\textbf{2.} Compute the VCRs, $\vf^t(\bx),\vw^t(\bx)$, for each supervision by Eq.~(\ref{eq:v1}) and (\ref{eq:v2}).}
  \State \multiline{\textbf{3.} Compute the normalized features $\tilde{f^t}(\bx)$.}
  \EndFor
  \State \multiline{Select the batch $\Df^+, \Dw^+$ using the VCRs, $\vf^t(\bx),\vw^t(\bx)$, and the normalized features $\tilde{f^t}(\bx)$ by Sec.~\ref{sec:batch}.}
\EndIf
  \State $\Df \leftarrow \Df \cup \Df^+, \Du \leftarrow \Du \setminus \Df^+$ 
  \State $\Dw \leftarrow \Dw \cup \Dw^+, \Du \leftarrow \Du \setminus \Dw^+$
  \State \multiline{Train $f^t$, $\hf^t$, and $\hw^t$ using $\Df$ and $\Dw$.}
\EndFor
\end{algorithmic}
\end{algorithm}
\subsection{Overview}
We propose an efficient algorithm to address the instance-wise supervision-level optimization problem within a fixed budget. 
Our approach comprises two primary components, as shown in Fig.~\ref{fig:overview}. 
The first component calculates the value-to-cost ratio (VCR) for each instance-supervision level pair, which indicates the expected improvement in model accuracy per unit of cost achieved through annotation for each instance. 
The VCR for each instance is determined by both the value of the supervision level and its uncertainty.
This allows us to evaluate the cost-effectiveness of assigning different supervision levels to each instance individually.
The second component performs batch selection within the budget $B$, optimizing the set of instances based on their VCRs and diversity. 
This two-part approach enables our approach to maximize the value of annotations within a limited budget by balancing cost-efficiency and diversity.

\subsection{Value-to-cost ratio (VCR) estimation \label{sec:VCR}}
In this section, we explain the calculation of the value-to-cost ratio (VCR) for each instance-supervision pair. An overview of the algorithm is provided in Algorithm~\ref{alg:overview}.
For an instance $\bx$ in the unlabeled pool $\Du$, the VCRs $\vf^t(\bx)$ for full supervision and $\vw^t(\bx)$ for weak supervision at round $t$ are determined by the uncertainty of the instance, the value, and cost for the two different-level labels, which are defined as:
\begin{align}
\vf^t(\bx) = \frac{\Mf^t \uf^t(\bx)}{\costf}  \label{eq:v1} \\
\vw^t(\bx) = \frac{\Mw^t \uw^t(\bx)}{\costw} 
\label{eq:v2}
\end{align}
where $\costf,\costw$ represent the annotation costs, and $\Mf^t,\Mw^t$ are the expected model improvement, where ${*}_{\rm f}, {*}_{\rm w}$ correspond to full supervision and weak supervision, respectively.
$\uf^t(\bx),\uw^t(\bx)$ represents the uncertainty for each instance $\bx$ in the unlabeled pool $\Du$ at round $t$, calculated based on model predictions.
This formulation allows us to evaluate the cost-effectiveness of assigning either full or weak supervision to each instance, providing a basis for the batch selection strategy within a fixed budget.

The expected model improvements $\Mf^t$ for full supervision and $\Mw^t$ for weak supervision (i.e., the values for each supervision) are estimated by evaluating the improvement in model performance as the training data is incrementally increased.
Consider the $t$-th annotation round, where fully supervised data $\Df^t$ and weakly supervised data $\Dw^t$ are available at that time.
To evaluate accuracy improvement, we divide the labeled data into $K$ subsets and incrementally add training data. For fully supervised data, the weakly supervised data is kept fixed while subsets of the fully supervised data are added. The model performance $m_{\mathrm{f},k}$ represents the evaluation using the first $k$ subsets of fully supervised training data, with $m_{\mathrm{f},1}$ being the performance when using just one subset. 
Similarly, we compute $m_{\mathrm{w},k}$ for the weakly supervised data, where the fully supervised data is fixed, and subsets of the weakly supervised data are incrementally added. 
Here, model performance is evaluated using a small validation set randomly sampled within the budget of a single round.
Then, a weighted average is applied for the improvement at each step, $m^t_{k+1}-m^t_{k}$ to smooth the sequence of improvements. The reason for the weighted approach is to focus on the latest model behavior.
To obtain the improvement per data point, each improvement value is normalized by dividing by $|\mathcal{D}|/K$, where $|\mathcal{D}|$ represents the number of instances.
That is, $\Mf^t$ and $\Mw^t$ are expressed as follows:
\begin{align}
\Mf^t = \frac{\sum_{k=1}^{K-1} k\cdot(m^t_{\mathrm{f}, k+1}-m^t_{\mathrm{f}, k})}{\sum_{k=1}^{K-1} k} \cdot \frac{K}{|\Df|} \label{eq:M1} \\
\Mw^t = \frac{\sum_{k=1}^{K-1} k\cdot(m^t_{\mathrm{w}, k+1}-m^t_{\mathrm{w}, k})}{\sum_{k=1}^{K-1} k} \cdot \frac{K}{|\Dw|}
\label{eq:M2}
\end{align}

The uncertainty $\uf^t(\bx)$ and $\uw^t(\bx)$, corresponding to full and weak supervision, can be flexibly defined based on the specific task and dataset.
For example, in the classification task, the entropy of the class prediction probabilities is adopted.
If different functions are used to calculate uncertainty for full and weak supervision, there is a risk that the scales of $\uf^t(\bx)$ and $\uw^t(\bx)$ may differ significantly. 
To address this, we normalize each uncertainty value by converting it to a percentile score. 
In this transformation, the instance with the highest uncertainty is assigned a score of $1$, the instance with the lowest uncertainty receives a score of $0$, and the instance ranked at the 10th percentile in terms of uncertainty is assigned a score of $0.9$.
We use percentiles rather than dividing by the maximum uncertainty value because extreme outliers could result in an overly large maximum, causing most values to be close to zero. This percentile-based normalization ensures that the distribution of uncertainty values remains well-scaled, preserving relative differences across instances without distortion due to outliers.

\subsection{Valuable and diverse batch selection using VCR \label{sec:batch}}

In this section, we describe the batch selection using the value-to-cost ratio (VCR) defined in the previous section.
The most straightforward approach for batch selection is to greedily select instances with the highest VCR values. 
However, as noted in previous studies (e.g.~\cite{sener2018active,ashdeep}), this approach can lead to selecting similar instances that have high VCR values, resulting in a loss of data diversity within the batch.

To address this, we propose a batch selection algorithm to select a diverse and valuable batch.
In other words, we select a group of instances with high VCR values and diverse features.
Specifically, we represent each instance as a vector using its VCR and feature vector. The vector differs depending on the type of label: $\vf(\bx)\tilde{f}(\bx)$ for full labels and $\vw(\bx)\tilde{f}(\bx)$ for weak labels. Here, $\tilde{f}(\bx)$ is a normalized feature vector (with a magnitude of 1) output by the feature extractor $f$. The optimal set of instances is selected based on the vectors.

We aim to select a set of instances that maximizes the area enclosed by the vertices of the selected vectors, as shown in the center of Figure~\ref{fig:overview}. Here, the optimization parameter is which vectors (instances) to select and their label types. When the selected instances have high diversity and value, the area enclosed by the selected vectors becomes larger.
We solve this optimization problem subject to budget constraints (i.e., the sum of the costs of the optimized supervision levels for instances is less than $B$).
The optimization problem of selecting vectors and determining their supervision levels can be reformulated as a selection problem, where the goal is to choose the optimal set of vectors from all possible vectors, including both those with weak labels $\{\vw(\bx)\tilde{f}(\bx)|\bx\in\Du\}$ and those with full labels $\{\vf(\bx)\tilde{f}(\bx)|\bx\in\Du\}$. This allows the problem of choosing the type of label to be integrated into the vector selection process.
This optimization problem can be solved by maximizing the determinant of a set of vectors subject to budget constraints. 

To perform the batch selection that maximizes the determinant of a matrix, one approach is to sample from a $k$-Determinantal Point Process ($k$-DPP)~\cite{kulesza2011k}.
This method selects a batch containing $k$ vectors with a probability proportional to the determinant of their Gram matrix. 
However, this differs from the batch selection in our approach in that we choose a set of vectors within a budget $B$ rather than choosing $k$ vectors.
Therefore, we propose a new selection method, which is inspired by the work~\cite{ashdeep} using the $k$-means++ seeding algorithm~\cite{arthur2007k} as an approximation algorithm for $k$-DPP. 

In our batch selection method, the vector of the batch is sampled sequentially with a probability proportional to the square of the distance to the vector closest to the one selected so far until the budget $B$ is exhausted.
Specifically, if the selected vector corresponds to full supervision (i.e., a vector in the set $\{\vf(\bx)\tilde{f}(\bx)|\bx\in\Du\}$, the cost $\costf$ is deducted from the budget, and it is added to the fully supervised data.
If it corresponds to weak supervision (i.e., a vector in the set $\{\vw(\bx)\tilde{f}(\bx)|\bx\in\Du\}$), the cost $\costw$ is deducted from the budget, and it is added to the weakly supervised data.
This process is repeated until the budget $B$ is exhausted.
By doing this, we can obtain a diverse and valuable batch.

\section{Experiments \label{sec:experiments}}
\subsection{Dataset}
To demonstrate the effectiveness of the proposed method, we utilized two datasets: CIFAR100 and Caltech-UCSD Birds-200-2011 (CUB200).
The task is a multi-class classification problem, where each class (full label) belongs to a superclass, and the superclass label is provided as a weak label.

CIFAR-100 is a dataset of natural images containing $100$ classes that span a wide range of everyday objects and animals. Each class in CIFAR-100 is associated with a predefined superclass, grouping similar classes together under broader categories, resulting in $20$ superclasses in total. The dataset consists of $60,000$ images in total, with $50,000$ images for training and $10,000$ images for testing

CUB200~\cite{WahCUB_200_2011} consists of $200$ classes of bird species, with each class representing a specific bird. 
Following previous study~\cite{Lu2018UsingCL}, we defined superclasses based on the suffixes in the class names, resulting in $70$ unique superclasses. 
The dataset contains $11,788$ images, with $5,994$ images designated for training and $5,794$ images for testing.

\subsection{Comparison methods}
We compared our approach with eight conventional AL methods, ranging from standard single-supervision AL methods to the state-of-the-art AL method incorporating multiple supervision, as follows:
\begin{itemize}
\item \textbf{Random:} The simplest baseline that samples instances randomly from the unlabeled pool. 
\end{itemize}
\begin{itemize}
\item \textbf{Margin~\cite{roth2006margin}:}  A method that selects instances in ascending order of the difference between the highest and second-highest predicted class probabilities, prioritizing instances with minimal margins.
\end{itemize}
\begin{itemize}
\item \textbf{MaxConf~\cite{wang2014new}:} An approach that selects instances in order of the lowest maximum predicted class probability, focusing on instances with the least confidence.
\end{itemize}
\begin{itemize}
\item \textbf{Entropy~\cite{wang2014new}:} A method that selects instances according to the highest entropy in the predicted class probabilities, targeting instances with the greatest uncertainty.
\end{itemize}
\begin{itemize}
\item \textbf{Coreset~\cite{sener2018active}:} A diversity-based sampling method that selects instances farthest from the labeled data using a greedy algorithm to enhance diversity. 
\end{itemize}
\begin{itemize}
\item \textbf{ALBL (Active Learning by Learning)~\cite{hsu2015active}:} A multi-armed bandit approach that selects between AL methods at each round. In this experiment, we utilized MaxConf and Coreset.
\end{itemize}
\begin{itemize}
\item \textbf{BADGE (Batch Active learning by Diverse Gradient Embeddings)~\cite{ashdeep}:} A method that considers both predictive uncertainty and diversity by selecting instances with diverse and high-magnitude gradients in the gradient space.
\end{itemize}
\begin{itemize}
\item \textbf{APFWA (Adaptive Proportion of Full and Weak Annotations)~\cite{tejero2023full}:} A state-of-the-art AL method that incorporates both full and weak supervision, which dynamically determines the proportion of full and weak annotations within each batch.
\end{itemize}

Since AL with both full and weak label annotation is a novel task, there is only one existing method for this task. Instead, we compare our approach with standard single-supervision AL methods as well as the state-of-the-art AL method. This comparison highlights the effectiveness of AL using weak labels. Additionally, a comparison with APFWA demonstrates the advantages of our approach for AL with both fully and weakly supervised labels. Furthermore, in Sec.~\ref{subsec:ALandWeak}, we also compare our approach with a simple combination of weak supervision and conventional AL methods to ensure a fair evaluation.

These methods offer a range of approaches, from simple random sampling to advanced techniques that balance uncertainty and diversity, as well as methods leveraging weak supervision, enabling a comprehensive evaluation of our proposed approach.

\subsection{Implementation details}
The network consists of a shared feature extractor $f$ and two classification heads, $\hf$ for full supervision and $\hw$ for weak supervision. The encoder of ResNet18~\cite{he2016deep} is used for the feature extractor $f$, and a single linear layer is used for each classification head.
For the network trained using both fully and weakly labeled data, we employ a two-stage training approach in which the feature extractor and each classification head are trained first with weakly supervised data and then with fully supervised data. Since both fully supervised and weakly supervised learning are classification tasks in this experiment, we used the cross-entropy loss. The learning rate was set to 0.001, and the optimizer was Adam~\cite{kingma2014adam}.
  
The settings for active learning were as follows. The number of rounds $T$ was set to $5$. The budget $B$ of each round was set to $1,000$ for CIFAR100 and to $500$ for CUB200 since CUB200 had approximately $5,000$ in training data, which was less than that of CIFAR100.
The validation data with full supervision were prepared from training data, $1,000$ for CIFAR1000 and $500$ for CUB200, which are the same number and small amount as in budget $B$.
The cost of full supervision $\costf$ was set to $1$, and the cost of weak supervision was set to $\frac{1}{2}$. In addition, in Sec.~\ref{subsec:cost}, we conducted the experiments when changing $\costw$ to $\frac{1}{4}$ and $\frac{1}{8}$ to investigate the impact of the difference on the cost of weak supervision.

The number of partitions $K$ for estimating the expected model improvements $\Mf^t$ and $\Mw^t$ was set to $5$.
As the performance metric of $m_{\mathrm{f},k}^t, m_{\mathrm{w},k}^t$, we used the classification accuracy. 
To enhance stability, we repeated the estimation process three times with different random seeds and calculated the average. 
For uncertainty measurement, we adopted the simplest approach, Margin~\cite{roth2006margin}, to evaluate instance uncertainty.

\section{Experimental results \label{sec:results}}
\subsection{Comparison with conventional active learning methods}
\begin{figure}[t]
  \begin{minipage}[b]{1.0\linewidth}
    \centering
    \includegraphics[keepaspectratio, scale=0.7]{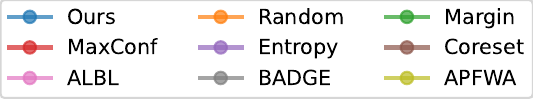}
    \vspace{2mm}
  \end{minipage}
  \begin{minipage}[b]{0.49\linewidth}
    \centering
    \includegraphics[keepaspectratio, scale=0.63]{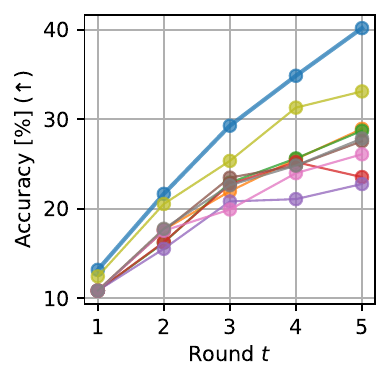}
    \subcaption{CIFAR100}
  \end{minipage}
  \begin{minipage}[b]{0.49\linewidth}
    \centering
    \includegraphics[keepaspectratio, scale=0.63]{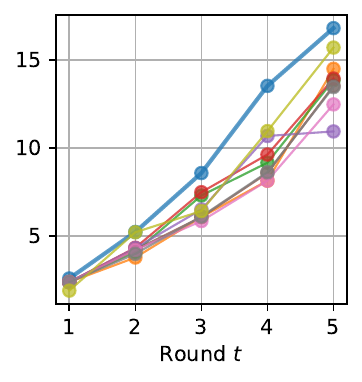}
    \subcaption{CUB200}
  \end{minipage}
  \vspace{-2mm}
  \caption{Comparison of our approach with the conventional active learning methods. The figures show the classification accuracy [\%] ($\uparrow$) for each round $t$ on CIFAR100 and CUB200. The cost of full supervision $\costf$ is $1$, and that of weak supervision is $\frac{1}{2}.$ \label{fig:comparison}}
\end{figure}

The comparison results with conventional AL methods on CIFAR100 and CUB200 are shown in Fig.~\ref{fig:comparison}(a), (b), respectively.
The graphs show the classification accuracy [\%] for each round $t$.

In Figure~\ref{fig:comparison}(a), the results on the CIFAR100 dataset demonstrate that our approach archives higher accuracy across all rounds compared to conventional methods, including random sampling, uncertainty-based AL method (Margin, MaxConf, Entropy), diversity-based AL method (Coreset), and other AL methods (ALBL, BADGE), which use only full supervision.
This demonstrates that our approach effectively leverages weak supervision, achieving higher performance within the same budget constraints.
For example, at round $t=5$, our approach achieved over 10\% higher accuracy compared to conventional AL methods that use only full supervision.
Furthermore, while these methods require five rounds (with a budget of 5000) to reach an accuracy of approximately 30\%, our approach achieved the same or even higher accuracy in just three rounds, using only three-fifths of the budget.

Our approach also surpasses APFWA, another AL approach that incorporates weak supervision by optimizing the proportions of weak and full supervision within the batch but does not focus on selecting instances. This result suggests that our approach more effectively selects cost-effective instances by dynamically considering instance-wise uncertainty and diversity, thereby maximizing the learning impact of weak supervision.

Similarly, Figure~\ref{fig:comparison}(b) presents the results on the CUB200 dataset, where our approach also outperforms other AL methods across all rounds.
The consistent improvement observed across both datasets indicates that our approach can dynamically optimize annotation strategies according to the dataset characteristics, achieving high adaptability.

\begin{figure}[t]
  \begin{minipage}[b]{1.0\linewidth}
    \centering
    \vspace{1mm}
    \includegraphics[keepaspectratio, scale=0.7]{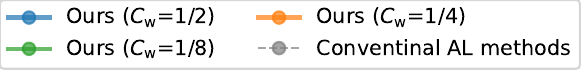}
  \end{minipage}
  \begin{minipage}[b]{0.49\linewidth}
    \centering
    \includegraphics[keepaspectratio, scale=0.63]{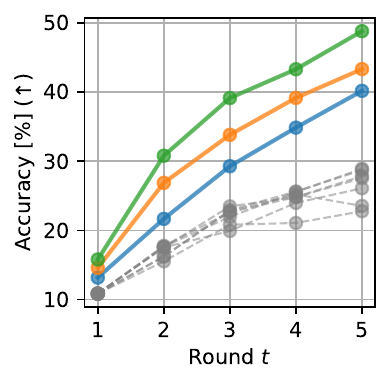}
    \subcaption{CIFAR100}
  \end{minipage}
  \begin{minipage}[b]{0.49\linewidth}
    \centering
    \includegraphics[keepaspectratio, scale=0.63]{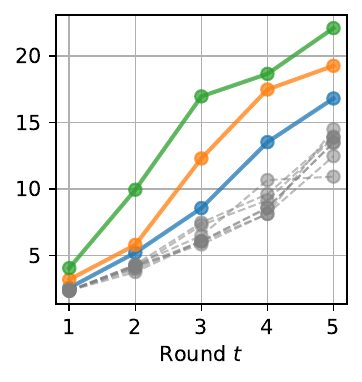}
    \subcaption{CUB200}
  \end{minipage}
  \vspace{-2mm}
  \caption{Classification accuracy [\%] ($\uparrow$) of the proposed method on CIFAR100 and CUB200 when the cost of weak supervision, $\costw$ is $\frac{1}{2}$, $\frac{1}{4}$, and $\frac{1}{8}$. Note that the cost $\costw$ is a value set by the user before active learning begins and is not a hyperparameter of the proposed method. This figure illustrates the effect of different weak supervision costs. For reference, we also include the results of conventional active learning methods (gray lines).}
  \label{fig:cost}
  \vspace{-2mm}
\end{figure}
\begin{figure*}[t]
  \begin{minipage}[b]{1.\linewidth}
    \centering
    \includegraphics[keepaspectratio, scale=0.4]{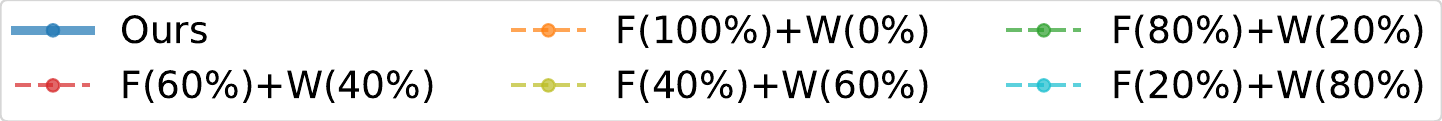}
  \end{minipage}
  \begin{minipage}[b]{0.33\linewidth}
    \centering
    \includegraphics[keepaspectratio, scale=0.63]{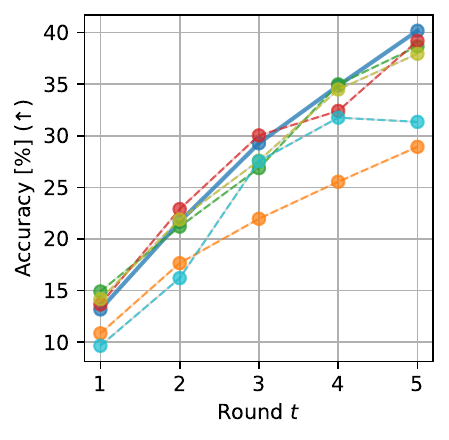}
    \subcaption{$\costw=1/2$}
  \end{minipage}
  \begin{minipage}[b]{0.33\linewidth}
    \centering
    \includegraphics[keepaspectratio, scale=0.63]{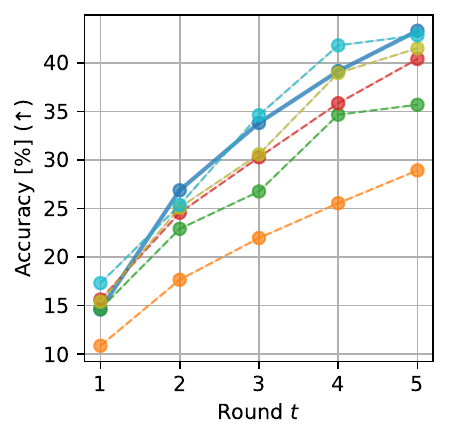}
    \subcaption{$\costw=1/4$}
  \end{minipage}
  \begin{minipage}[b]{0.33\linewidth}
    \centering
    \includegraphics[keepaspectratio, scale=0.63]{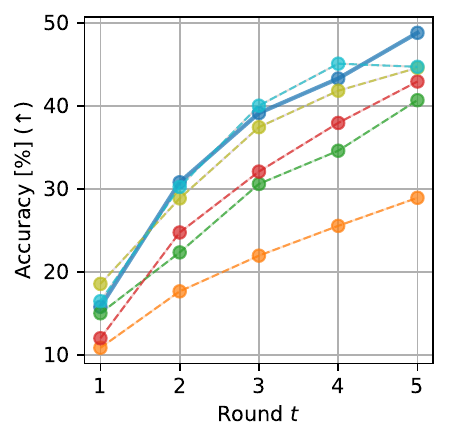}
    \subcaption{$\costw=1/8$}
  \end{minipage}
  \caption{Comparison with combined active learning and multiple supervision baseline. The classification accuracy [\%] ($\uparrow$) on CIFAR100 with fixed proportions of full and weak supervision within each batch. Each batch was sampled with fixed proportions, and conventional active learning methods were applied separately for full and weak supervision.}
  \label{fig:fixed_ratio}
\end{figure*}
\subsection{Analysis for weak supervision costs \label{subsec:cost}}
To demonstrate that our approach works at any cost of weak supervision and that the cheaper the cost of weak supervision, the greater the effect, we conducted experiments when changing the cost of weak supervision.
Figure~\ref{fig:cost} shows the classification accuracy [\%] on CIFAR100 and CUB200 at different costs of weak supervision.
In this experiment, the cost of full supervision $\costf$ is fixed at $1$, meaning that $\costw$ represents the relative cost to full supervision.
It is important to note that the cost $\costw$ is defined before active learning begins and is not a hyperparameter of the proposed method.
Imagine a scenario where the fee paid to a non-expert annotator, who can only assign rough class labels (weak labels), is lower than that of an expert annotator.
The cost $\costw$ varies for each task or dataset, corresponding to the rewards offered by crowdsourcing, etc.

In previous sections, we set the cost of weak supervision to $\costw=\frac{1}{2}$.
Here, we investigate the effects of our approach when further reducing $\costw$ to $\frac{1}{4}$ and $\frac{1}{8}$.
As shown in Figure~\ref{fig:cost}, for both datasets, the classification accuracy improves as the cost of weak supervision decreases.
This improvement is due to the fact that the lower cost of weak supervision allows more data to be annotated within the same budget, thereby enhancing the training process. In other words, the cheaper the cost of weak supervision, the more impactful weak supervision and our approach become.

\subsection{Comparison with combined active learning and multiple supervision baseline}
\label{subsec:ALandWeak}
In addition to comparing our approach with conventional active learning approaches, we conduct experiments to evaluate our approach against a straightforward combination of weak supervision and conventional active learning techniques. This baseline approach applies a fixed proportion of weak supervision within the budget, utilizing conventional active learning strategies within each supervision type.
\footnote{APFWA~\cite{tejero2023full} is a method that dynamically determines the proportion of full and weak annotations. 
It must use random instance sampling to estimate the optimal proportion, and as a result, it cannot be combined with the conventional instance-wise AL methods (e.g., margin-based sampling).}

Specifically, in this combined method, weak supervision is allocated at a constant rate—20\%, 40\%, 60\%, and 80\% in the batch. Within each full and weak supervision, conventional AL methods are employed to select data. In this experiment, the active learning method for each full or weak supervision employed was sampling based on margin, which is used for the uncertainty of the proposed method.

Figure~\ref{fig:fixed_ratio} shows the comparison between our approach and baseline methods that combine active learning with weak supervision on the CIFAR100 dataset under different costs for weak supervision ($\costw=\frac{1}{2}$,$\frac{1}{4}$, and $\frac{1}{8}$).
In each case, the baseline methods use fixed proportions of full and weak supervision within each batch, denoted as F($x\%$)+W($y\%$), where $x$ and $y$ represent the percentages of full and weak supervision, respectively.

Across all weak supervision costs, our approach consistently achieves high accuracy.
An important point to note is that the optimal proportion of full and weak supervision is unknown beforehand and varies depending on the cost of weak supervision.
For example, when the cost of weak supervision $\costw$ is $\frac{1}{2}$, the configuration with 60\% full supervision and 40\% weak supervision achieves the best accuracy among the baseline methods. 
However, when the cost $\costw$ is reduced to $\frac{1}{8}$, a configuration with 20\% full supervision and 80\% weak supervision performs best.
In contrast, our approach dynamically constructs optimal batches by optimizing both the supervision level and instance selection based on the cost-effectiveness obtained from the trained model at each round.

\begin{figure}[t]
  \begin{minipage}[b]{1.0\linewidth}
    \centering
    \includegraphics[keepaspectratio, scale=0.7]{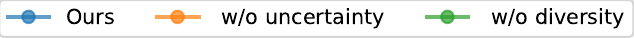}
  \end{minipage}
  \begin{minipage}[b]{0.49\linewidth}
    \centering
    \includegraphics[keepaspectratio, scale=0.63]{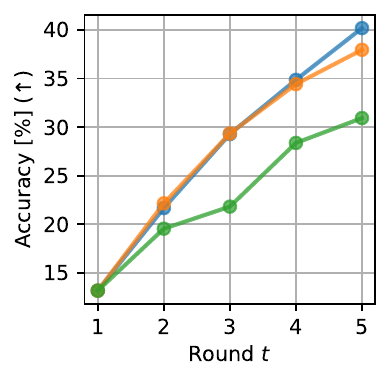}
    \subcaption{CIFAR100}
  \end{minipage}
  \begin{minipage}[b]{0.49\linewidth}
    \centering
    \includegraphics[keepaspectratio, scale=0.63]{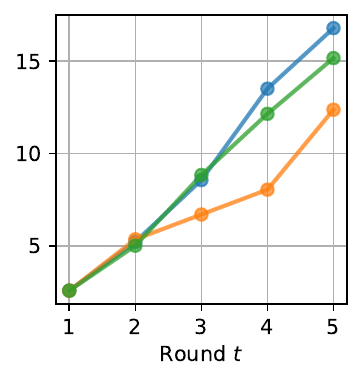}
    \subcaption{CUB200}
  \end{minipage}
  \vspace{-2mm}
  \caption{Ablation study comparing the proposed method without instance-wise uncertainty $\uf(\bx), \uw(\bx)$ (Ours w/o uncertainty) and without batch selection considering diversity (Ours w/o diversity). The figures show the classification accuracy [\%] ($\uparrow$) for each round $t$ on CIFAR100 and CUB200.}
  \label{fig:ablation}
\end{figure}
\subsection{Ablation study}
We conducted ablation studies to demonstrate the effectiveness of instance-wise value (i.e., uncertainty) and the batch selection with diversity on CIFAR100 and CUB200, as shown in Fig.~\ref{fig:ablation}.

The first ablation variant, referred to as ``Ours w/o uncertainty,'' removes the instance-wise uncertainty component from the VCR calculation. 
In this variant, the VCR is determined solely based on the annotation cost and expected model improvement without taking into account the uncertainty of each instance. Consequently, the supervision level is not optimized for each instance individually. 
The results show that our approach outperforms this variant on both datasets.
This effect is particularly prominent in the CUB200 dataset, demonstrating the benefit of optimizing ``instance-wise'' supervision-level.

The second ablation variant, called `Ours w/o diversity,'' involves a greedy approach to batch selection, where instances are selected in descending order of VCR without considering data diversity. 
Our approach, which incorporates diversity into the batch selection process, consistently outperforms this variant on both datasets. 
The impact of diversity-aware selection is especially evident in the CIFAR100 dataset, demonstrating that considering diversity in batch selection significantly enhances model performance.

\section{Conclusion, limitation, and future work \label{sec:conclusion}}

In this paper, we proposed a novel active learning (AL) approach, called Instance-wise Supervision-level Optimization (ISO), which optimizes the supervision level (either full or weak annotation) for each instance to maximize annotation efficiency within a budget. By evaluating the uncertainty, diversity, and value-to-cost ratio (VCR) of each instance for full and weak supervision, our approach automatically determines the instances to be annotated and their optimal annotation level. Experiments on classification tasks demonstrated that ISO outperforms not only standard AL methods but also state-of-the-art AL methods with weak supervision, while effectively leveraging instance-specific information to improve both accuracy and cost-efficiency.
\par

A current limitation of this study is that the effectiveness of our ISO has been validated only in classification experiments. The concept of ISO is applicable to various tasks beyond classification. For example, ISO can be applied to object segmentation tasks, where precise yet costly object boundaries are assumed for full supervision, while simpler rectangular bounding boxes are used for weak supervision. Another current limitation is the assumption that supervision levels are only two. We can expand to deal with more supervision levels by introducing class levels other than exact classes and superclasses. In future work, we aim to demonstrate the full versatility of our approach by exploring these extensions across a broader range of tasks, reinforcing its potential as a scalable, cost-effective solution for real-world applications.


\vspace{17pt}
\noindent
{\bf Acknowledgements}: This work was supported by JSPS KAKENHI Grant Number JP23KJ1723, and JST, ACT-X Grant Number JPMJAX23CR, Japan.
\clearpage

{
    \small
    \bibliographystyle{ieeenat_fullname}
    \bibliography{main}
}


\end{document}

%% file: preamble.tex
\usepackage{bm}
\newcommand{\bx}{{\bm x}}

\newcommand{\uf}{u_\mathrm{f}}
\newcommand{\uw}{u_\mathrm{w}}
\newcommand{\Mf}{M_\mathrm{f}}
\newcommand{\Mw}{M_\mathrm{w}}
\newcommand{\costf}{C_\mathrm{f}}
\newcommand{\costw}{C_\mathrm{w}}
\newcommand{\vf}{v_\mathrm{f}}
\newcommand{\vw}{v_\mathrm{w}}
\newcommand{\hf}{h_\mathrm{f}}
\newcommand{\hw}{h_\mathrm{w}}

\newcommand{\Df}{\mathcal{D}_\mathrm{f}}
\newcommand{\Dw}{\mathcal{D}_\mathrm{w}}
\newcommand{\Du}{\mathcal{D}_\mathrm{u}}